\documentclass[journal]{IEEEtran}
\usepackage{flushend}
\usepackage[american]{babel}
\usepackage{subcaption}
\usepackage{booktabs}
\usepackage{bm}
\usepackage{graphicx}
\usepackage{amsmath,amssymb}
\usepackage{algorithm,algorithmic}
\usepackage{tikz}

\newcommand{\hl}[1]{\textcolor{black}{#1}}

\usepackage[breaklinks=true,colorlinks,bookmarks=false]{hyperref}

\begin{document}

\title{Spatio-Temporal Point Process for\\ Multiple Object Tracking}

\author{Tao Wang, Kean Chen, Weiyao Lin, John See, Zenghui Zhang, Qian Xu, and Xia Jia%
\thanks{This paper is supported in part by the following grants: China Major Project for New Generation of AI Grant (No. 2018AAA0100400), National Natural Science Foundation of China (No. 61971277), ZTE Research Grant, and Shanghai ’The Belt and Road’ Young Scholar Exchange Grant (No. 17510740100).}
\thanks{T. Wang, K. Chen, W. Lin, Z. Zhang are with the Department of Electronic Engineering, Shanghai Jiao Tong University, China (email: \{wang\_tao1111,ckadashuaige,wylin,zenghui.zhang\}@sjtu.edu.cn).}
\thanks{J. See is with the Faculty of Computing and Informatics, Multimedia University, Malaysia (email: johnsee@mmu.edu.my).}
\thanks{Q. Xu and X. Jia are with ZTE Corporation, China (email: \{xu.qian5,jia.xia\}@zte.com.cn.}
\thanks{Corresponding author: Weiyao Lin.}
}

\maketitle

\begin{abstract}
Multiple Object Tracking (MOT) focuses on modeling the relationship of detected objects among consecutive frames and merge them into different trajectories. MOT remains a challenging task as noisy and confusing detection results often hinder the final performance. Furthermore, most existing research are focusing on improving detection algorithms and association strategies. As such, we propose a novel framework that can effectively predict and mask-out the noisy and confusing detection results before associating the objects into trajectories. In particular, we formulate such ``bad'' detection results as a sequence of events and adopt the \textit{spatio-temporal point process} to model such events. Traditionally, the occurrence rate in a point process is characterized by an explicitly defined intensity function, which depends on the prior knowledge of some specific tasks. Thus, designing a proper model is expensive and time-consuming, with also limited ability to generalize well. To tackle this problem, we adopt the convolutional recurrent neural network (conv-RNN) to instantiate the point process, where its intensity function is automatically modeled by the training data. Furthermore, we show that our method captures both temporal and spatial evolution, which is essential in modeling events for MOT. Experimental results demonstrate notable improvements in addressing noisy and confusing detection results in MOT datasets. An improved state-of-the-art performance is achieved by incorporating our baseline MOT algorithm with the spatio-temporal point process model.
\end{abstract}


\section{Introduction}

\begin{figure}[t]
\small
\centering
\includegraphics[width=1\linewidth]{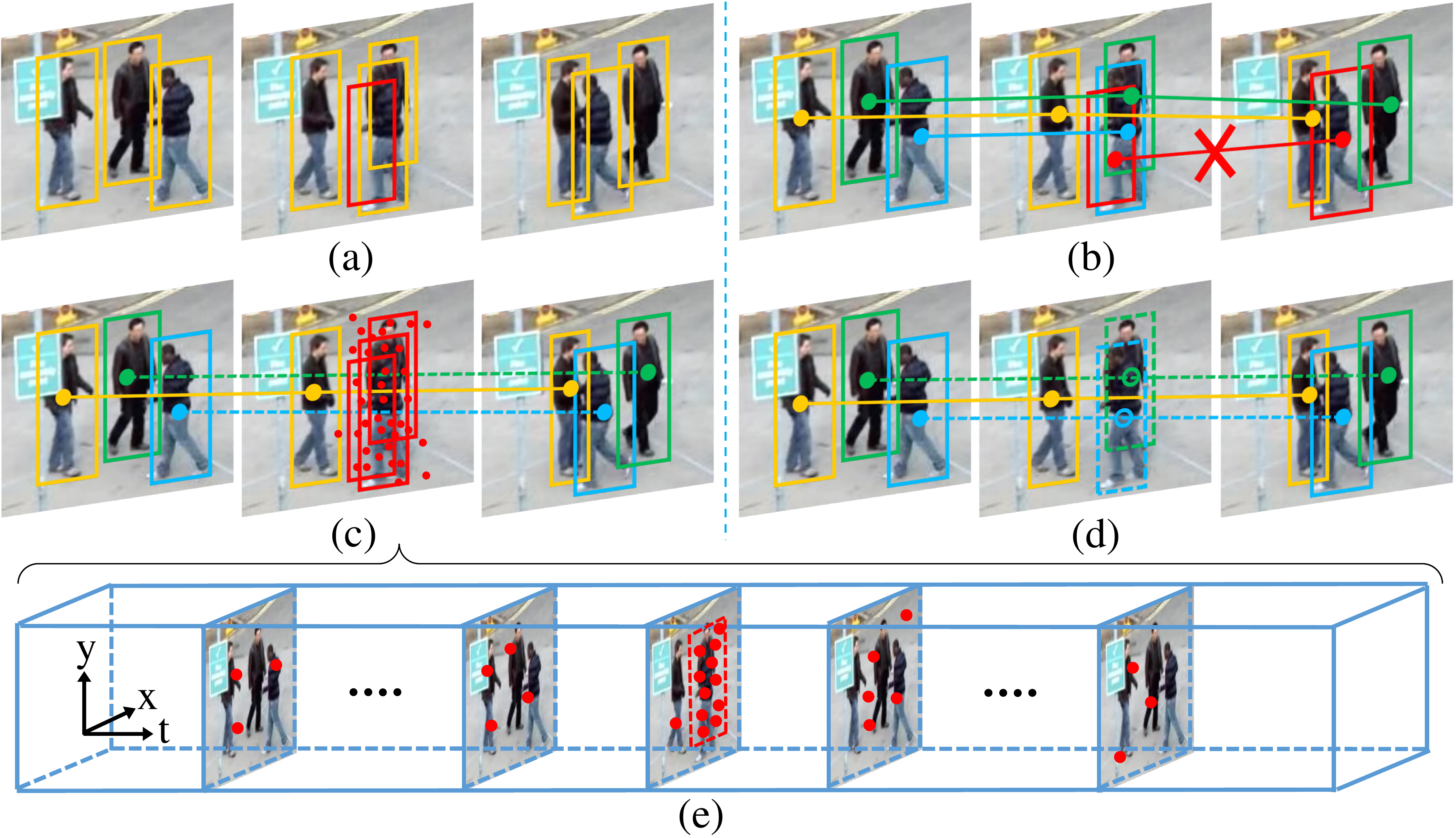}
\vspace{-4.5mm}
  \caption{An example of a noisy object detection. (a) The detection results, where a potential noisy detection (red box) occurs in the scene. (b) As a result, the red trajectory is unnecessarily tracked. This is an incorrect result. (c) Our approach can predict the area which is likely to contain the noisy detections, and avoid them in the association process. (d) The tracking result using our method, where the dashed line boxes are generated by linear interpolation. (e) The prediction is generated by the proposed point process model.}
\vspace{-1.0ex}
\label{fig-1}
\end{figure}

\begin{figure}[t]
\small
\centering
\includegraphics[width=1\linewidth]{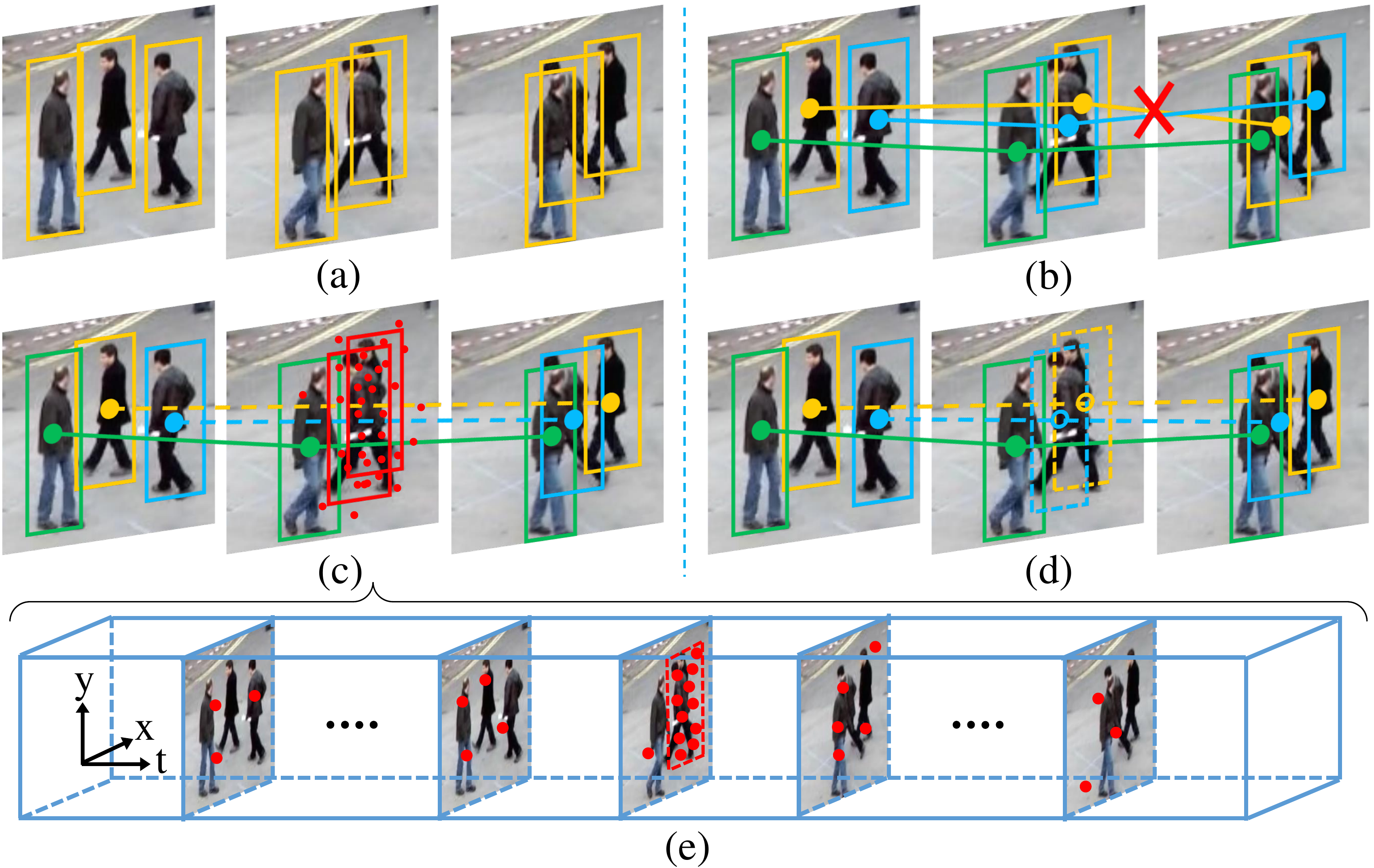}
\vspace{-4.5mm}
  \caption{An example of a confusing object detection. (a) The detection results. (b) An incorrect tracking result due to objects that are overlapping in close proximity (yellow box and blue box). (c) Our approach can predict the area which is likely to contain the confusing objects, and avoid them in the association process. (d) The tracking result by our method, where the dashed line boxes are generated by linear interpolation. (e) The prediction is generated by the proposed point process model.}
\vspace{-1.0ex}
\label{fig-2}
\end{figure}

Multiple object tracking (MOT) is one of the fundamental problems in computer vision, which is important in many applications like intelligent video surveillance, behavior analysis, automatic driving and robotics. MOT constitutes the task of modeling the relationship of detected objects among consecutive frames and then merging them into different trajectories~\cite{tang2015subgraph,schulter2017deep,yang2017hybrid}. This task remains challenging, one major issue is that some ``bad'' detection results always hinder the performance of MOT. \hl{In general, such ``bad'' results can be divided into two different types: 1) noisy detection results, \textit{i.e.} false positives in object detection, and 2) confusing detection results, \textit{i.e.} two highly overlapping objects with similar appearances.} Two such examples are shown in \autoref{fig-1} and \autoref{fig-2}. It can be seen that, both the noisy and confusing objects had misguided the matching process. Furthermore, we observe that most of the failure examples in MOT are caused by them, directly or indirectly. Hence, the main problem in MOT that needs to be addressed, can realistically be reduced to: \textit{how to effectively handle these ``bad'' detection results?}

Some existing methods in MOT such as \cite{tang2015subgraph,yang2017hybrid,berclaz2011multiple,gao2017beyond,shi2015detection,yu2007multiple,benfold2011stable,emami2018machine,jin2012hierarchical} improved the tracking performance by introducing more robust object association strategies. More specifically, some advanced techniques for cost function and corresponding optimization algorithm have been developed to associate objects in different frames. However, these methods do not explicitly model the ``bad'' detection results, so they can be confused by the objects with high similarity in appearance and motion. Other methods including \cite{wang2016joint,bae2017confidence,xiang2018multiple,fang2018recurrent,zhou2018online,hu2017moving} focused their attention on different feature representations and metrics for detected objects. These methods have better accuracy in normal scenes but are still affected by the noisy detection results. Some works like \cite{henschel2018fusion,yu2016poi,chu2017online,he2017sot,chen2017enhancing,taalimi2015robust} adopted more accurate object detectors in attempt to reduce these noisy detections, while the performances are still hindered by confusing detections such as the highly overlapping objects.

To tackle this issue in MOT, we propose a framework that can effectively predict and mask-out these ``bad'' detection results before associating the objects into trajectories. First, we note that the ``bad'' detections can be formulated as a sequence of events that happen in different frames and locations. Thus, we need to infer when and where these events are likely to happen, given the feature of current frame and the historical behaviour of the detector as prior information. More specifically, we model these events based on the pixels that are inside the bounding boxes of ``bad'' detections. Such events are distributed across the spatio-temporal tube of a video, and are generated based on complicated latent mechanisms~\cite{lin2016tube,peng2018tracklet}, which is hard to capture through simple modeling. \hl{Because these events happen in the motion of objects, so there exists relationship between these events which actually reflects the motion information of the objects. In other words, we can detect these bad detections more accurately and improve the performance of MOT by obtaining the relationship between these events. Moreover, we make the following observations: (1) Noisy object detections are more likely to appear in the area where there are already some noisy detections in the previous frames. (2) If there are confusing detections among people in a group who walk closely or dress similarly to others, confusing detections are more likely to appear in these people in the subsequent frames.} In this paper, we introduce the use of \textit{spatio-temporal point process} to deal with such events.

Point process~\cite{daley2007introduction} is a powerful tool for modeling the real-world sequential data, which has lots of applications in many fields, such as finance~\cite{bacry2015hawkes}, equipment maintenance~\cite{yan2013towards,ertekin2015reactive} and social network~\cite{du2015dirichlet,farajtabar2016multistage}. A point process is characterized by its conditional intensity function, which presents the occurrence rates of some class of events conditioned on the historical events. Traditionally, the intensity function can be explicitly defined based on prior knowledge of event data and latent mechanisms of the process. The intensity function usually consists of two parts~\cite{xiao2019learning}: an exogenous intensity that describes factors driven by the inherent and often time-varying occurrence rate for a type of events; an endogenous intensity which describes the triggering effect from previous events. This parametric strategy has been widely adopted in many classic models, such as Poisson Processes~\cite{kingman2005p}, Hawkes processes~\cite{hawkes1971spectra} and self-correcting processes~\cite{isham1979self}. However, there are three issues that needs to be handled in the case of MOT:
\textbf{(1)} Designing a proper parametric model is expensive and time-consuming, since it requires expert domain knowledge and experience if it were to be modeled manually. Besides, the generalization ability is also limited.
\textbf{(2)} It is difficult to properly capture the dynamics of influence from historical information given the complicated nature of data patterns in MOT task. On the other hand, it is also hard for traditional point processes to incorporate other heterogeneous data, such as time series~\cite{xiao2019learning} associated with event sequences.
\textbf{(3)} Furthermore, the detection events in MOT occur in different frames and different locations in the scene, which requires capturing both temporal and spatial evolution.

To tackle these problems, we adopt the \textit{spatio-temporal point process} to model the detection events, where a convolutional recurrent neural network (conv-RNN) is proposed to instantiate the point process. We note that: \textbf{(1)} In our method, the intensity function is automatically modeled by the training data, without requiring expert knowledge and experience. \textbf{(2)} We propose a two-stream RNN framework to handle two different inputs, \textit{i.e.} time series and event sequence, with a novel time evolving mechanism to align and merge these two input data. This enables our model to capture the complex dynamics of influence from historical information. \textbf{(3)} We incorporate the use of 
convolution operation in RNN, which enables spatial diffusion for the historical information of events. Coupled with the capacity of RNN in modeling temporal dependence, the proposed conv-RNN based point process is equipped with the capability of capturing both temporal and spatial evolution. The main contributions of this paper are summarized as follows:

\begin{itemize}
\vspace{-0mm}
\setlength{\topsep}{0pt}
\setlength{\itemsep}{1pt}
\setlength{\parskip}{1pt}
\item We propose a novel framework in MOT that can effectively predict and mask-out the noisy and confusing detection results before associating the objects into trajectories. The ``bad'' detection results are formulated as a sequence of events and are modeled by \textit{spatio-temporal point process}.
\item We introduce a two-stream pipeline to handle two synchronous and asynchronous inputs (\textit{i.e.} time series and event sequence) with a novel evolving mechanism for merging.
\item We propose a convolutional based recurrent neural network to instantiate the point process, where the temporal and spatial evolution are well captured.
\item We show notable performance improvement by incorporating the proposed method into state-of-the-art MOT algorithms across different metrics and benchmark datasets.
\end{itemize}

\begin{figure*}[t]
\small
\centering
\includegraphics[width=1\linewidth]{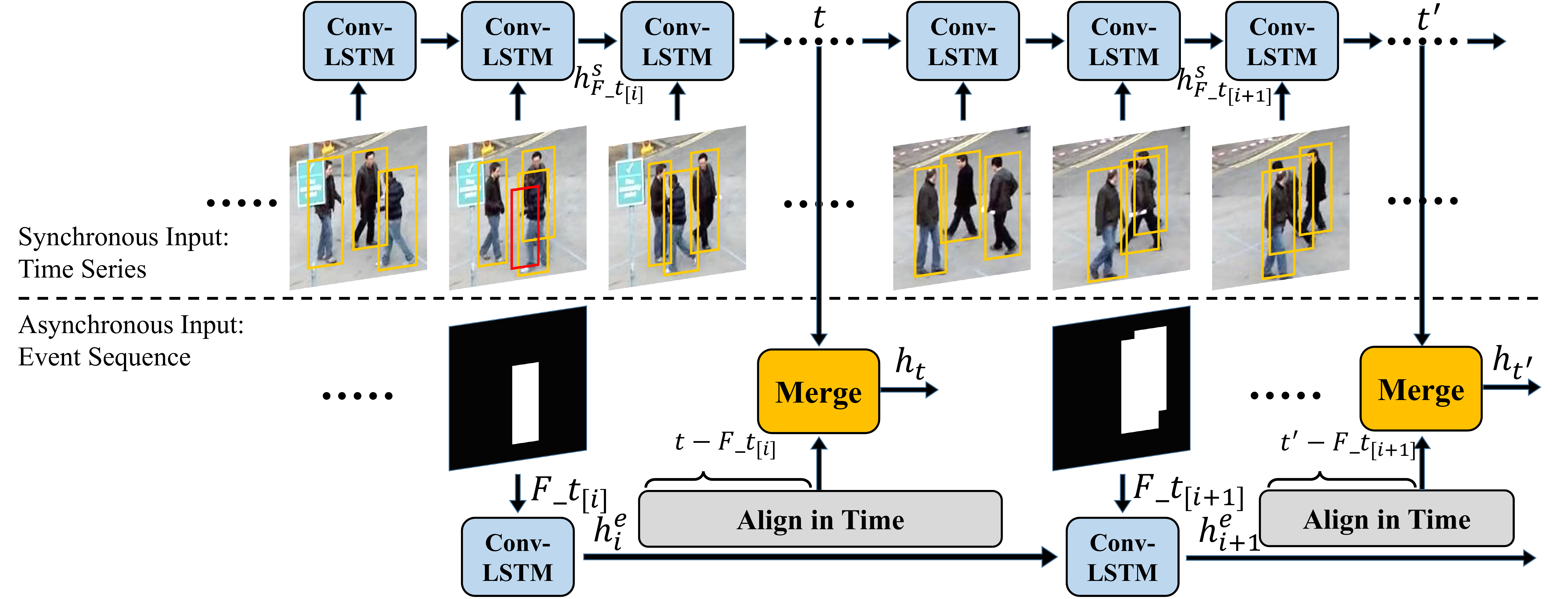}
\vspace{-4.5mm}
  \caption{The framework of proposed point process model. There are two RNNs in our model. The first one is for modeling the time series, which represents the background information and features at each time. In our approach, the features and background information are the raw image and the detection results in this frame. The second one is for modeling the event sequence, which represents the historical events and their time-stamps. In our approach, the events are defined to be the pixels which are inside the ``bad'' detection results. The feature \(h_i^e\) which encodes the event sequence is aligned in time and synchronized with the feature \(h_t^s\) from time series. \([i]\) denotes the \(i\)-th equivalence class of events that have same time-stamp, so that \(t_{[i]}\) means the \(i\)-th frame that contains at least one event.}
\vspace{-1.0ex}
\label{fig-3}
\end{figure*}

\section{Related Works}
\subsection{Multiple Object Tracking}
Multiple object tracking (MOT) is widely applied in many fields like intelligent video surveillance, behavior analysis, automatic driving and robotics~\cite{bae2017confidence,choi2015near}. Multiple object tracking involves the process of modeling the relationship of detected objects among consecutive frames and then merging them into different trajectories~\cite{tang2015subgraph,schulter2017deep,yang2017hybrid}. This task remains challenging due to several problems. An obvious observation is that certain ``bad'' detection results remain the bottleneck for better performance in MOT. Generally, these ``bad'' results can be divided into two different types, reflecting two main causes: 1) \emph{noisy} detection results, \textit{i.e.} false positives in object detection, 2) \emph{confusing} detection results, \textit{i.e.} two highly overlapping objects with similar appearances. Among existing works, there are three known aspects of how these issues have been handled:
\textbf{(1)} Some existing methods in MOT such as \cite{tang2015subgraph,yang2017hybrid,berclaz2011multiple,gao2017beyond,shi2015detection,yu2007multiple,benfold2011stable,emami2018machine} improve the tracking performance by introducing more robust object association strategies. For example, \cite{berclaz2011multiple} proposed a constrained flow optimization problem to model the objects association, and use k-shortest path algorithm~\cite{yen1971finding} to solve this problem. \cite{yang2017hybrid} modeled the data association into a min-cost multi-commodity network flow which fuses both global optimization and local optimization.
\textbf{(2)} Other methods such as \cite{wang2016joint,bae2017confidence,xiang2018multiple,fang2018recurrent,zhou2018online} focus their attention towards studying different feature representations and metrics in attempt to mitigate these errors. In addition, \cite{gao2018osmo} proposed an attention-based appearance model to obtain a better similarity metric.
\textbf{(3)} Some research works \cite{henschel2018fusion,yu2016poi,chu2017online,he2017sot,chen2017enhancing,taalimi2015robust} adopted more accurate object detectors to reduce these problems, particularly noisy detections. For instance, \cite{henschel2018fusion} proposed a multi-detector to track pedestrian through fusing the detection of head and body. \cite{yu2016poi} proposed a high-performance detector for MOT which combines skip pooling~\cite{bell2016inside} and multi-region strategies~\cite{gidaris2015object}. Although these methods give valuable results in MOT, their performances are still limited due to intrinsic limitations; in particular, these methods contained unaddressed deficiencies. Firstly, these methods do not explicitly model ``bad'' detection results. As such, incorrect tracking caused by confusion of objects with high similarity in appearance and motion cannot be learned or modeled. Secondly, these methods demonstrated good tracking performance in normal scenes whereas in scenes with noisy detection results, they remain highly susceptible to erroneous tracking.
However, the proposed approach explicitly models the dynamic and relationship of the ``bad'' detections, and is able to predict and mask-out them before associating the objects into trajectories.

\subsection{Spatio-Temporal Point Processes}
Temporal point process has been widely applied in many fields such as finance~\cite{bacry2015hawkes}, equipment maintenance~\cite{yan2013towards,ertekin2015reactive} and social network~\cite{du2015dirichlet,farajtabar2016multistage}. The point process is characterized by the conditional intensity function, which presents the occurrence rates of events conditioned on the historical data. Traditionally, the intensity function is explicitly defined based on prior or expert knowledge about event data and latent mechanisms of the process. However, in many tasks, it is hard to design a parametric model by hand. Recently, several research works~\cite{du2016recurrent,xiao2019learning} studied the use of a neural network based model for temporal point process. \cite{du2016recurrent} proposed a semi-parametric model for point process where the events are treated as a univariate point process and the event types are treated as marks that are associated with events. Their key idea is to view the intensity function of a temporal point process as a nonlinear function of the event history, and use a recurrent neural network to automatically learn a representation of influences from the event history. \cite{xiao2019learning} introduced recurrent point process networks which instantiate temporal point process models with temporal recurrent neural networks (RNNs). Their key idea is to model the intensity function by two RNNs: one temporal RNN captures the relationships among events, while the other RNN updates the intensity functions based on time series. On the other hand, the spatio-temporal point process~\cite{gonzalez2016spatio} reveals two main pieces of information (space and time) about the data considered. The data is treated as the realisation of a random collection of points, which evolves in space and time. It can be viewed as a temporal point process with further (spatial) dimensions.

Spatio-temporal point process also has wide applications such as earthquakes~\cite{marsan2008extending} and disease outbreaks~\cite{choi2015constructing}. For the case of the MOT task, this is potentially feasible since we are also interested in both the spatial and temporal evolutions of detected objects. In contrast to the aforementioned existing schemes which define a neural network model for temporal point process, we propose convolutional recurrent neural networks (conv-RNN) to instantiate the point process. The convolution operation in RNN enables spatial diffusion for the historical information of events. Combined with the capacity of RNN in modeling temporal dependencies, our conv-RNN based point process can effectively capture both temporal and spatial evolutions.

\subsection{Recurrent Neural Networks}
Recurrent Neural Network (RNN)~\cite{werbos1988generalization,elman1990finding,pascanu2013difficulty} is a kind of neural network tailored for modeling sequences such as time series data. RNN allows connections among hidden units to be associated with a time delay, which is useful to encode historical information and capture the relationship between events that evolve over time. However, a vanilla RNN model is difficult to train and does not handle long-range dynamics well due to the \textit{vanishing gradient} and \textit{exploding gradient} problems~\cite{bengio1994learning}. To this end, the Long Short-Term Memory (LSTM) is proposed, which has shown to be stable and powerful for modeling long-range dependencies. The key idea of LSTM lies in the memory cell \(c_t\), which acts as an accumulator of the state information. It allows gradient to be trapped in the cell and be prevented from vanishing too quickly~\cite{xingjian2015convolutional}. Traditional LSTM has 1-D vectors for the cell input, output and hidden state. Such LSTMs contain too much redundancy for spatial data. To address this problem, \cite{xingjian2015convolutional} proposed Conv-LSTM which has convolutional structures in both the input-to-state and state-to-state transitions. Further on, by stacking multiple Conv-LSTM layers, we can build a network model that is feasible for spatio-temporal sequence forecasting problems such as the MOT task. In this paper, we design a variant of the described Conv-RNN model, which can handle different inputs for both long and short term information, and can incorporate the spatio-temporal varying components to build the intensity. The proposed method will be elaborated in detail in Section \ref{sec:method} after describing the MOT task in Section \ref{sec:prelim}.

\section{Preliminaries}
\label{sec:prelim}
\subsection{Problem Definition}
The input data is a sequence \(C=\{I_1, I_2, ...\}\) of consecutive frames in a video, combined with the object detection results \(D_j=\{d^j_1, d^j_2, ...\}\) in each frame \(I_j\). Note that the detection results \(D_j\) are generated by some baseline detection algorithms and are not guaranteed to be entirely correct. For example, the false positive detections that always occur in crowded scenes are regarded as noisy detections. Besides, other erroneous detection results are caused by confusion from detected objects of similar appearances which overlapped with each other during motion. The MOT task aims to model the relationship among these detected objects and associate them into trajectories. Hence these noisy or confusing detection results will hinder the performance of MOT. In this paper, we adopt the spatio-temporal point process to address this issue.

Let \(B_j=\{b^j_1, b^j_2, ...\}\) denote the set of ``bad'' detections in each frame \(I_j\) (note that \(B_j \subseteq D_j\)). In our spatio-temporal point process, we define the events \(e\) to be the pixels \(e=(t, x, y)\) that are inside some ``bad'' detection results \(b^t_i\) in frame \(I_t\). We use \(\mathcal{E}=\{e_1, e_2, ..., e_n\}\) to denote the set of events from all frames. In general, the dynamics of spatio-temporal point process is captured by its conditional intensity function \(\lambda^* (t,x,y)\) (where \(*\) emphasizes that this function is conditional on the history), which can be defined as:
\begin{equation}
\lambda^*(t',x',y')=\frac{\partial^3 E[N(t,x,y)|\mathcal{H}_t]}{\partial t\partial x\partial y}|_{t=t',x=x',y=y'}
\end{equation}
where \(\mathcal{H}_t\) denotes the history of events before time \(t\) and \(N(t,x,y)\) is the counting function which counts the number of events with spatial coordinates less than \((x,y)\) and temporal coordinates less than \(t\), \hl{while \(E[N]\) is the expectation of the counting function}. Since we assume that the point process satisfies the regularity condition:
\begin{equation}
P\{N([t,t+\epsilon_1],[x,x+\epsilon_2],[y,y+\epsilon_3])>1\}=o(\epsilon_1\epsilon_2\epsilon_3)
\label{eq:regularity}
\end{equation}
\hl{where \(P\) measures the probability of event (its argument), \({[t,t+\epsilon_1],[x,x+\epsilon_2],[y,y+\epsilon_3]}\) indicates a small area around the point \((t, x, y)\) while \(o(\epsilon_1\epsilon_2\epsilon_3)\) refers to a term of higher-order infinitesimal \textit{w.r.t} \(\epsilon_1\epsilon_2\epsilon_3\). Equation~\ref{eq:regularity} shows that in point process there will be at most one event within a short time and in a small area.} The intensity function \(\lambda^*(t,x,y)\) can also be interpreted as the conditional probability density that an event occurs at \((t,x,y)\). Thus the probability density for an event sequence \(\{e_1,e_2,...\}\) can be written as:
\begin{equation}
\label{eq2}
f(e_1,e_2,...)=\prod_j f^*(t_j,x_j,y_j|\mathcal{H}_{t_j})
\end{equation}
where
\begin{equation}
\label{eq1}
f^*(t,x,y|\mathcal{H}_{t_j})=\frac{\lambda^*(t,x,y)}{ \exp(\int_{\widetilde{t_{j}}}^{t}\int_{u,v}\lambda^*(\tau,u,v)d\tau du dv)}
\end{equation}
Here, \(\widetilde{t_j}\) denotes the maximal time-stamp \(t_k\) that satisfies \(t_k < t_j\), the denominator presents the probability that no new event occurs up to time \(t_j\) since \(\widetilde{t_{j}}\).
Point processes are always learned by maximizing the likelihood function; whereby its implementation is not specified.
Thus, one problem is how to define and implement the intensity function \(\lambda^*(t,x,y)\) in the context of MOT.
\subsection{Tracklet-Based Algorithm}
\label{sec:tracklet}
\begin{figure}[t]
\small
\centering
\includegraphics[width=1\linewidth]{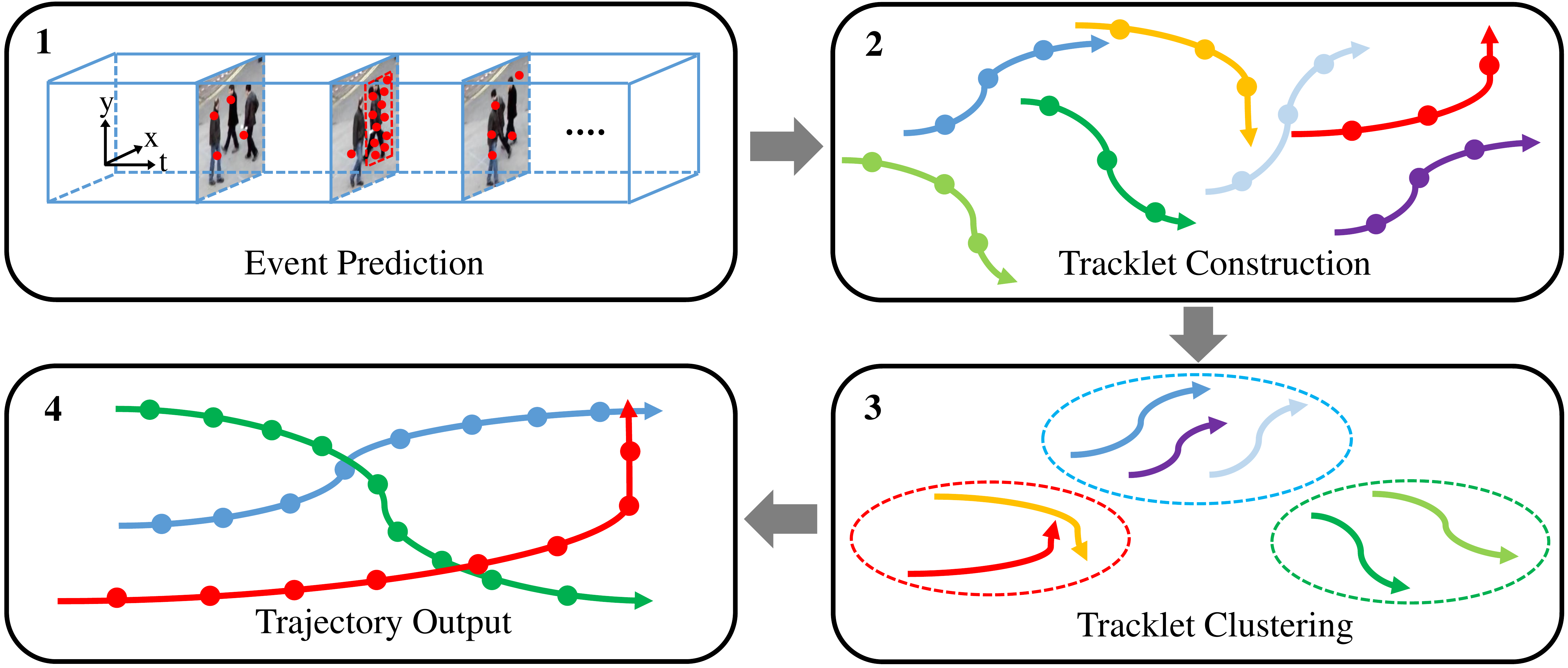}
\vspace{-4.5mm}
  \caption{Framework of the tracking method. First, the spatio-temporal point process is adopted to predict ``bad'' detection events. \hl{The ``bad'' detection boxes, which cover too many events, will be directly deleted from each frame.} Then, the detections (dots) are linked to form different tracklets (arrows). Finally, the tracklets are clustered and associated to generate complete trajectories.}
\vspace{-1.0ex}
\label{fig-4}
\end{figure}

In this paper, our baseline tracking method follows the tracklet-based strategy~\cite{peng2018tracklet,wang2016joint,taalimi2015robust} to associate objects from each frame into trajectories. Similar to \cite{peng2018tracklet}, the baseline tracking method contains three steps. The first step is \textit{tracklet construction}, which aims to merge highly related objects into high-confidence tracklets. Note that we take a strict threshold of appearance similarity and motion similarity for object matching. This ensures that the generated tracklets are highly reliable and consistent. Then, these tracklets are treated as basic units for the subsequent clustering step. The second step is \textit{tracklet similarity calculation}. In some works~\cite{tang2015subgraph,yang2017hybrid}, tracklet similarity is calculated directly from features of the terminal object (\textit{i.e.} object on the last frame of the tracklet), which cannot capture information of the entire tracklet. Instead of using only the terminal object, we determine tracklet similarity by using all the appearance information in the tracklets. More details can be found in \cite{peng2018tracklet}. The third step is \textit{tracklet clustering}. After obtaining the tracklets and corresponding similarities, the clustering process associates these tracklets into complete trajectories. The basic idea in~\cite{peng2018tracklet} is that cluster centers are characterized by a higher local density than their neighbors and by a relatively larger distance from other sample points with higher local densities. The local density of \(i\)-th tracklet is defined as:
\begin{equation}
\rho_i=\sum_{j: O(i,j)=0} H(s_{ij}-s_c)
\end{equation}
where \(s_c\) is a threshold which is always set to 0.5, \(H(x)\) is the Heaviside step function, \(s_{ij}\) denotes the similarities between \(i\)-th and \(j\)-th tracklets. The constraint \(O(i,j)=0\) ensure that these two tracklets have no overlap in the temporal domain. The maximum similarity of \(i\)-th tracklet is defined as:
\begin{equation}
\delta_i=\max_{j:\rho_j>\rho_i,O(i,j)=0}s_{ij}
\end{equation}
Then, the tracklets that have a maximum similarity \(\delta_i<s_c\) are marked as cluster centers such that the similarities between any two cluster centers are always lower than \(s_c\). Finally, the remaining tracklets are assigned to the same cluster according to the most similar tracklet of higher density. 

	\subsection{Combination of Tracking and Point Process}
	The whole proposed tracking framework is shown in \autoref{fig-4}. At first, we use the spatio-temporal point process to model intensity function \(\lambda^*(t,x,y)\)
	for each frame \(I_j\) of the sequence \(C\). Then we predict ``bad'' detection events for each frame and remove those detection bounding boxes which contain events more than a given threshold. After this step, we get better detection results \(D^*_j\) for each frame of sequence \(C\). Then \(D^*_j\) will be used as the input of our tracking baseline to construct short tracklets. Finally, the tracklets are clustered and associated to generate complete trajectories.
	\par
    \hl{
    It should be noted that although an event pixel can be contained by more than a single box, the definition of events in our method still matches well with that in point process. In our method, ``bad'' events are not ``bad'' detection boxes but the ``bad'' pixels in detection boxes. So, if a ``bad'' event on a pixel is contained by many bounding boxes, it still represents \emph{one} event, not many events. Furthermore, when an event happens in the overlapped area of two bounding boxes, it also means that this event decreases the confidence of both boxes.
    }
\section{Method}
\label{sec:method}
In this section, we describe in detail the method which characterizes the proposed neural network model for spatio-temporal point process.

As shown in \autoref{fig-3}, we have two RNNs in the proposed model. The first one, a synchronous network, is for modeling the time series~\cite{xiao2019learning}, which represents the background information and features at each time step. We use the raw image and the detection results in each frame to model the features and background information. The second one, an asynchronous network, is for modeling the event sequence, which represents the historical events and their timestamps. \hl{``Bad'' detections are defined in two ways: (1) noisy detections, which are bounding boxes that are false positives or inaccurate detections, (2) confusing detections, which are bounding boxes that are easily mistaken for other objects. The labeling strategy for the ``bad'' detection is shown in Section \ref{section_experiments}.} The events are defined to be the pixels which are inside the ``bad'' detection results. We take the binary map as the input to each recurrent unit (LSTM), where the pixels are masked to 1 if the event occurs. The feature \(h_i^e\) which encodes the event sequence is aligned and synchronized with the feature \(h_t^s\) from time series. Then, we merge the features from two RNNs and calculate the intensity function based on it. After obtaining the intensity function, we can infer the event through sampling on the probability density function (\textit{c.f.} \autoref{eq1}). In the testing phase, for each bounding box \(b\), we calculate the ratio \(r_b\) between the number of events \hl{in it} and its area. Then, a detection \(b\) is treated as a ``bad'' detection if \(r_b\) is larger than some threshold.

\subsection{Synchronous RNN}
The synchronous RNN takes in time series as input data. We have two kinds of time series: one is the raw images; one is the binary maps for detection results, where the pixels that are inside a detected bounding box are masked to 1, otherwise 0. As shown in \cite{khoreva2017simple}, though the mask of bounding box is a weaker label than the segmentation mask of the object, it still provides sufficient information to train the segmentation task. However, we intend for the proposed model to focus on learning potential areas of ``bad'' detections instead of actual segmentation of objects. Thus, the binary maps of bounding boxes are used as auxiliary input data instead of training label.

At each step \(t\), we first send the raw image \(I_t\) and the binary map \(M^D_t\) of detection results to a CNN (\textit{e.g.} VGG, ResNet) to extract feature maps. Then the feature maps are merged and fed to the recurrent unit (conv-LSTM) to extract temporal features. Therefore, the hidden states \(h_t^s\) for the time series can be obtained by:
\begin{equation}
(h_t^s,c_t^s)=ConvLSTM(h_{t-1}^s,c_{t-1}^s,\psi_1(I_t,M^D_t))
\end{equation}
where \(\psi_1\) denotes the feature extractor CNN, \(c_t^s\) denotes the cell status and both \(h_t^s,c_t^s\) are feature maps.
Note that such input data is sampled uniformly over time, which can reflect the exogenous intensity in point process.

\subsection{Asynchronous RNN}
The asynchronous RNN aims to capture the relation between events. It takes in an event sequence as input data. In this spatio-temporal point process for MOT, each event provides both spatial and temporal features. For spatial feature, our strategy is similar to that in previous section. We use binary map to represent the spatial information of events that occur in a frame, \textit{i.e.}, those pixels that are inside the ``bad'' detections are masked to 1, otherwise 0. For the temporal feature, we use the inter-event duration \(F\_t_{[i]}-F\_t_{[i-1]}\) for each event occurring at time \(F\_t_{[i]}\). Here, \([i]\) denotes the \(i\)-th
equivalent class of events, so \(F\_t_{[i]}\) denotes the \(i\)-th frame that contains at least one event.
We reformulate the scalar \(F\_t_{[i]}-F\_t_{[i-1]}\) to a map that has same size as the raw image and all pixels are represented by this scalar. Then, these two maps are concatenated and passed through a CNN feature extractor and conv-LSTM to obtain the hidden state \(h_{i}^e\):
\begin{equation}
h_{i}^e=ConvLSTM(h_{i-1}^e,c_{i-1}^e,\psi_2(M^B_i,F\_t_{[i]}-F\_t_{[i-1]}))
\end{equation}
where \(\psi_2\) denotes the CNN feature extractor, \(M^B_i\) denotes the binary map of ``bad'' detection results, \(c_i^e\) denotes the cell status and both \(h_i^e,c_i^e\) are feature maps.
Note that the events are always sparsely located and unevenly distributed in the time domain, which shows that the sequences are asynchronous in nature and the temporal intervals can be very long. Thus, one advantage of such asynchronous RNN is that it can better capture long-term dependencies of events, which can reflect the endogenous intensity in point process.

\subsection{Align and Merge}
\label{sec:evolve}
We aim to construct the intensity function based on the features from both time series and event sequence. To this end, the asynchronous features should be first aligned and synchronized with the time series. Traditionally, one may consider multiplying a decaying function \(\gamma(t-F\_t_{[i]})\) with the feature \(h_i^e\) to simulate the influence from historical events to current time, where \(\gamma(t)\) can be specified by \(\exp(-t)\). However, the latent dynamics of point process in MOT are still unknown. Directly specifying the evolving function \(\gamma(t)\) in such a manner may lead to the model misspecification problem~\cite{du2016recurrent}. Therefore, we use a 
\hl{three-layer MLP whose input is the concatenation of \(t-F\_t_{[i]}\) and \(h_i^e\)} to learn the evolving function \(\psi_3(\cdot,t)\), where the model capacity can be sufficiently large to cover any dynamic patterns. Assuming current time is \(t\) and the latest hidden state of the event \(e\) is \(h_i^e\), we have:
\begin{equation}
\hat{h}^e_t=\psi_3(h_i^e,t-F\_t_{[i]})
\end{equation}
where \(\hat{h}^e_t\) denotes the aligned event feature at time \(t\). After that, we merge the outputs from both time series and event sequence to obtain \(h_t=[h^s_t,\hat{h}^e_t]\).
Finally, the intensity function \(\lambda(t)\) can be formulated as
\begin{equation}
\label{eq3}
\lambda^*(t)=\sigma(w^s h^s_t+w^e \hat{h}^e_t)
\end{equation}
where \(\lambda^*(t)\) in this equation is an intensity map, the \(w^s h^s_t\) term reflects the exogenous intensity, the \(w^e \hat{h}^e_t\) term reflects the endogenous intensity, and \(\sigma\) denotes the activation function. A reasonable choice of the activation function would be one that ensures that the intensity function is non-negative and almost linear when the input is much greater than 0. Several activation functions fit this purpose: the \emph{softplus} function, defined as \(log(exp(x)+1)\) or exponential linear unit \emph{(elu)}, defined as \(elu(x)+1\) where \(elu(x)=x\) when \(x\geq 0\) and \(elu(x)=\exp(x)-1\) when \(x<0\).

\subsection{Learning Method}
Given the sequence of events \(E=\{e_1,e_2,\ldots,e_n\}\), we obtain the intensity function \(\lambda^*(t,x,y)\) by \autoref{eq3}. Then, the log-likelihood function can be formulated based on \autoref{eq2} and \autoref{eq1} as follows:
\begin{equation}
\label{eq4}
\small
\log f=\sum_j \log\lambda^*(t_j,x_j,y_j)-\int_{t_0}^{t_n}\int_{u,v}\lambda^*(\tau,u,v)d\tau dudv
\end{equation}
The first part is the accumulative summation of log-intensity function for those events occurring at \(\{(t_1,x_1,y_1),(t_2,x_2,y_2),\ldots,(t_n,x_n,y_n)\}\). The second part is the integral of intensity over space and time where no event happens. In our implementation, we \hl{set the spatio-temporal resolution to be 1 and} adopt a discrete approximation of \autoref{eq4}:
\begin{equation}
\small
\log f=\sum_j \log\lambda^*(t_j,x_j,y_j)-\sum_{(\tau,u,v)\neq (t_j,x_j,y_j)}\lambda^*(\tau,u,v)
\end{equation}
This objective function is differentiable and can be efficiently optimized by stochastic gradient descent.
\hl
{
\subsection{Testing Method}
In testing phase, for each video, we put the detection results of each frame and output of previous frames into the proposed point process model. Through this way, we obtain an intensity map for each frame \(t\) which is then used to infer the event by sampling on the probability density function. For each bounding box \(b_i\) in frame \(t\), we count the number of events in it and calculate the ratio \(r_i\) between event number and box area. If \(r_i\) is larger than a given threshold we will consider \(b_i\) as a ``bad'' detection and remove it from frame \(t\).
After repeating the above process for each frame of the testing video, all ``bad'' detections would have been removed and we proceed to use the remaining detection results as the input of the tracking method to form tracklets.
}
\section{Experiments}\label{section_experiments}
\subsection{Experimental Settings}
\subsubsection{Datasets}
Our experiments are performed on two benchmark datasets: MOT16~\cite{milan2016mot16} and MOT17. MOT16 dataset contains 7 training and 7 testing sequences. MOT17 has the same video sequences as MOT16, with three different detection sets (DPM~\cite{felzenszwalb2009object}, Faster-RCNN~\cite{ren2015faster}, and SDP~\cite{yang2016exploit}). The video sequences are captured by both static and moving cameras, with different scenes and resolutions. The viewpoint can also vary significantly from each other, \textit{e.g.}, the camera may be overlooking the scene from a high position, a medium position (at pedestrian's height), or at a low position (at ground level). Also, various forms of object occlusion and large variation in object appearances render these datasets challenging.

\subsubsection{Evaluation Metrics}
In MOT Challenge
Benchmark~\cite{leal2015motchallenge,milan2016mot16}, the relationship between ground truth and tracker output is established using bounding boxes. The intersection over union (IOU) is used as the similarity criterion, where the threshold is set to 0.5. Then, the tracking performance is measured by Multiple Object Tracking Accuracy (MOTA, the primary metric), Multiple Object Tracking Precision (MOTP), the total number of False Negatives (FN), the total number of False Positives (FP), the total number of Identity Switches (IDs), the percentage of Mostly Tracked Trajectories (MT) and the percentage of Mostly Lost Trajectories (ML). Specifically, MOTA measures the overall tracking performance of an approach, combined with FN, FP and IDs. Furthermore, AP (Average Precision) is adopted to directly measure the prediction accuracy of events.
\hl{
\subsubsection{Sub-network Implementation}
The proposed model is implemented as follows. We use Resnet-50 as the backbone network and construct a Siamese network to extract the features of detected boxes. For the Conv-LSTM, the number of LSTM units is set to 8 and the hidden unit size is 1024. Meanwhile, we use a three-layer MLP to model the evolving function whose input is the concatenation of \(t-F\_t_{[i]}\) and \(h_i^e\).
}
\hl{
\subsubsection{``Bad'' Detections labeling}
As for noisy object detections, we observe every detection box in the MOT dataset and label the boxes whose IoU (with any ground truth) are smaller than 0.5 as noisy detections. We then run the baseline algorithm on the dataset and label the mismatched detections as confusing object detections. With the above method, we can locate most of the ``bad'' detections.
}

\subsubsection{Parameter Settings}
In our experiments, we use the public detection results provided by the MOT16 and MOT17 datasets, so as to have a fair comparison with other MOT
methods. In the ablation study (Section \ref{sec:ablation}), we use four training sequences of MOT16 as training data and the other three training sequences for validation. We use the tracklet-based algorithm introduced in Section \ref{sec:tracklet} as our baseline algorithm for MOT. We also choose deep-SORT~\cite{wojke2017simple} and Tracktor~\cite{bergmann2019tracking} as other baselines for comparison purposes. To train the network, the standard Adam optimizer is chosen with the batch size of 32 and initial learning rate of 0.001. After every 40,000 iterations, the learning rate is further decayed by 10\% of the initial learning rate. The training process terminates after 150,000 iterations.

\subsection{Ablation Study and Discussion}
\label{sec:ablation}
\subsubsection{Comparison on Different Settings}
\begin{table}[t]
\small
\centering
\caption{Ablation experiments on the evolving function, activation function and training loss. Models are tested on MOT2016 dataset.}
\begin{subtable}[t]{1\linewidth}
\centering
\begin{tabular}{cccccccc}
\midrule[1pt]
Evolving Function & MOTA\(\uparrow\) & MOTP\(\uparrow\) & MT\(\uparrow\) & ML\(\downarrow\) \\
\midrule[1pt]
Parametric Decaying & 42.3 & 77.4 & 18.8\% & 40.8\% \\
Neural Network & \textbf{44.1} & \textbf{82.7} & \textbf{23.8\%} & \textbf{36.0\%} \\
\midrule[1pt]
\end{tabular}
\label{evolving_function}
\caption{}
\end{subtable}
\quad
\begin{subtable}[t]{1\linewidth}
\centering
\begin{tabular}{cccccccc}
\midrule[1pt]
Activation \(\sigma(\cdot)\) & MOTA\(\uparrow\) & MOTP\(\uparrow\) & MT\(\uparrow\) & ML\(\downarrow\) \\
\midrule[1pt]
Sigmoid & 41.9 & 77.5 & 18.6\% & 42.2\% \\
Biased Relu & 42.4 & 77.2 & 19.0\% & 40.0\% \\
Elu & 43.1 & 79.5 & 20.0\% & 37.7\% \\
Softplus & \textbf{44.1} & \textbf{82.7} & \textbf{23.8\%} & \textbf{36.0\%} \\
\midrule[1pt]
\end{tabular}
\label{activation_function}
\caption{}
\end{subtable}
\quad
\begin{subtable}[t]{1\linewidth}
\centering
\begin{tabular}{cccccccc}
\midrule[1pt]
Training Loss & MOTA\(\uparrow\) & MOTP\(\uparrow\) & MT\(\uparrow\) & ML\(\downarrow\) \\
\midrule[1pt]
Cross Entropy & 40.5 & 77.8 & 16.2\% & 48.0\% \\
Mean Squared Loss & 42.7 & 77.3 & 19.1\% & 40.8\% \\
Log-likelihood & \textbf{44.1} & \textbf{82.7} & \textbf{23.8\%} & \textbf{36.0\%} \\
\midrule[1pt]
\end{tabular}
\label{training_loss}
\caption{}
\end{subtable}
\label{parameter_settings}
\end{table}
\begin{table}[t]
	\small
	\centering
	\caption{Ablation experiments on different components. AP is adopted to measure the prediction accuracy.}
	\begin{tabular}{cccc}
		\midrule[1pt]
		&Time Independent & Sync RNN & Sync/Async RNN \\
		\midrule[1pt]
		AP & 26.9\% & 27.3\% & \textbf{35.4}\% \\
		Speed/FPS & \textbf{61} & \hl{43} & \hl{37} \\
		\midrule[1pt]
	\end{tabular}
	\label{ablation_study_ap}
\end{table}
In this section, we investigate the impact of some practical modifications introduced in Section \ref{sec:method}. All results are shown in \autoref{parameter_settings}.

\noindent\textbf{Evolving Function:}
First, we study the evolving function (\textit{c.f} Section \ref{sec:evolve}), and report tracking performances based on different choice of implementation. Results indicate that the neural network model clearly outperforms the parametric decaying function. This verifies our previous hypothesis that the model capacity of neural network is much larger than the simple parametric evolving function. This helps the model to capture more complex dynamic patterns. Therefore, the evolving function is implemented using neural network for the subsequent studies.

\noindent\textbf{Activation Function:}
Secondly, we study the activation function (\textit{c.f.} Section \ref{sec:evolve}), and report tracking performances using different \(\sigma\) values. As explained earlier, we argue that the intensity function should be non-negative and almost linear when the input is much greater than 0. Thus, the observation that the sigmoid function has the worst result validates our point. Note that the biased Relu function (\(\epsilon+\max\{x,0\}\)) satisfies these conditions but performs much worser than softplus function. This is likely because the gradient of the Relu function is strictly eliminated when \(x\) is smaller than 0. Thus, the model also trained non-optimally when the biased Relu is adopted as the activation function.

\noindent\textbf{Training Loss:}
Thirdly, we study the impact of training loss in the optimization of the network and report the results. We observe marginal benefits using log-likelihood function over mean squared error function and cross-entropy loss function. So, we adopt the log-likelihood as our choice of objective function in all the following studies.

\subsubsection{Comparison on Different Components}

\begin{table}[t]
\small
\centering
\caption{Ablation experiments on different components. All models are tested on MOT2016 dataset.}
\begin{tabular}{cccccccc}
\midrule[1pt]
Components & MOTA\(\uparrow\) & MOTP\(\uparrow\) & MT\(\uparrow\) & ML\(\downarrow\) \\
\midrule[1pt]
Baseline & 38.2 & 77.7 & 16.1\% & 49.3\% \\
+ Time Independent & 39.6 & 77.6 & 16.6\% & 49.7\% \\
+ Sync RNN & 41.6 & 77.1 & 19.0\% & 39.3\% \\
+ Sync/Async RNN & \textbf{44.1} & \textbf{82.7} & \textbf{23.8\%} & \textbf{36.0\%} \\
\midrule[1pt]
Deep-SORT~\cite{wojke2017simple} & 28.4 & \textbf{78.8} & 4.8\% & 63.1\% \\
+ Time Independent & 29.1 & 78.5 & 5.8\% & 59.4\% \\
+ Sync RNN & 29.9 & 78.4 & 6.6\% & 57.4\% \\
+ Sync/Async RNN & \textbf{30.4} & 78.2 & \textbf{7.2\%} & \textbf{55.1\%} \\
\midrule[1pt]
\hl
{Tracktor~\cite{bergmann2019tracking}} & 42.4 & 78.2 & 17.3\% & 40.4\% \\
+ Time Independent & 42.9 & 78.0 & 17.7\% & 39.9\% \\
+ Sync RNN & 43.9 & 78.3 & 18.5\% & 39.1\% \\
+ Sync/Async RNN & \textbf{44.5} & \textbf{78.5} & \textbf{19.2\%} & \textbf{38.6\%} \\
\midrule[1pt]
\end{tabular}
\label{ablation_study}
\end{table}

\begin{figure}[t]
\small
\centering
\includegraphics[width=1\linewidth]{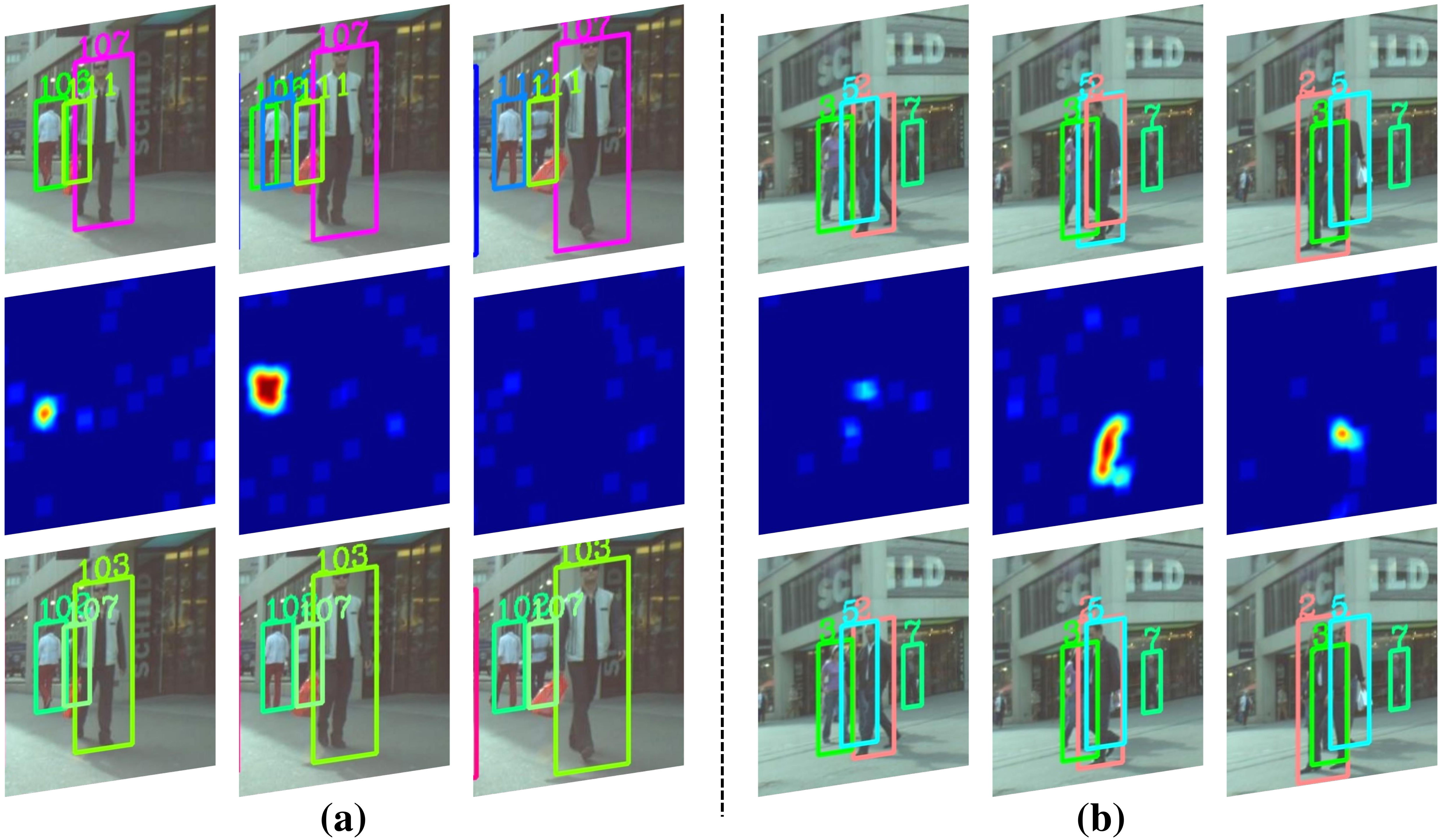}
\vspace{-4.5mm}
  \caption{Two tracking examples. First row: the results by baseline tracking algorithm. Second row: the intensity maps generated from the proposed point process model. Third row: the results by our proposed method. (a) An example of noisy detection. The noisy detection occurs in the second frame of the top row (green box), which causes two different trajectories for one object. (b) An example of confusing detection. The confusing detections occurs in the second frame of the top row (blue and red boxes), which causes a mismatch error (identity switch).}
\vspace{-1.0ex}
\label{fig-5}
\end{figure}

\begin{table*}[t]
\footnotesize
\centering
\caption{Tracking performances of our approach and state-of-the-art methods on MOT2016.}
\setlength{\tabcolsep}{6mm}{
\resizebox{1\linewidth}{!}{
\begin{tabular}{cccccccc}
\midrule[0.5pt]
Method & MOTA\(\uparrow\) & MOTP\(\uparrow\) & MT\(\uparrow\) & ML\(\downarrow\) & FP\(\downarrow\) & FN\(\downarrow\) & IDS\(\downarrow\) \\
\midrule[0.5pt]
OICF~\cite{kieritz2016online} & 43.2 & 74.3 & 11.3\% & 48.5\% & 6651 & 96515 & 381 \\
CBDA~\cite{bae2017confidence} & 43.9 & 74.7 & 10.7\% & 44.4\% & 6450 & 95175 & 676 \\
QCNN~\cite{son2017multi} & 44.1 & 76.4 & 14.6\% & 44.9\% & 6388 & 94775 & 745 \\
STAM~\cite{chu2017online} & 46.0 & 74.9 & 14.6\% & 43.6\% & 6895 & 91117 & 473 \\
MDM~\cite{tang2016multi} & 46.3 & 75.7 & 15.5\% & 39.7\% & 6449 & 90713 & 663 \\
NOMT~\cite{choi2015near} & 46.4 & 76.6 & 18.3\% & 41.4\% & 9753 & 87565 & \textbf{359} \\
JGD-NL~\cite{levinkov2017joint} & 47.6 & 78.5 & 17.0\% & 40.4\% & \textbf{5844} & 89093 & 629 \\
TSN-CC~\cite{peng2018tracklet} & 48.2 & 75.0 & \textbf{19.9\%} & \textbf{38.9\%} & 8447 & 85315 & 665 \\
LM-PR~\cite{tang2017multiple} & 48.8 & \textbf{79.0} & 18.2\% & 40.1\% & 6654 & 86245 & 481 \\
\midrule[0.5pt]
{\bf Ours} & \textbf{50.5} & 74.9 & 19.6\% & 39.4\% & 5939 & \textbf{83694} & 638\\
\midrule[0.5pt]
\end{tabular}}}
\label{state-of-the-art-mot16}
\end{table*}

\begin{table*}[t]
\footnotesize
\centering
\caption{Tracking performances of our approach and state-of-the-art methods on MOT2017.}
\setlength{\tabcolsep}{6mm}{
\resizebox{1\linewidth}{!}{
\begin{tabular}{cccccccc}
\midrule[0.5pt]
Method & MOTA\(\uparrow\) & MOTP\(\uparrow\) & MT\(\uparrow\) & ML\(\downarrow\) & FP\(\downarrow\) & FN\(\downarrow\) & IDS\(\downarrow\) \\
\midrule[0.5pt]
EEBMM~\cite{maksai2019eliminating} & 44.2 & - & - & - & - & - & \textbf{1529} \\
NG-bL~\cite{kim2018multi} & 47.5 & \textbf{77.5} & 18.2\% & 41.7\% & 25981 & 268042 & 2069 \\
OGSDL~\cite{fu2018particle} & 48.0 & 77.2 & 17.1\% & 35.6\% & 23199 & 265954 & 3998 \\
DMAN~\cite{zhu2018online} & 48.2 & 75.9 & 19.3\% & 38.3\% & 26218 & 263608 & 2194 \\
EDM~\cite{chen2017enhancing} & 50.0 & 77.3 & 21.6\% & 36.3\% & 32279 & 247297 & 2264 \\
MHT~\cite{kim2015multiple} & 50.7 & \textbf{77.5} & 20.8\% & 36.9\% & 22875 & 252889 & 2314 \\
DLCS~\cite{long2018real} & 50.9 & 76.6 & 17.5\% & 35.7\% & 24069 & 250768 & 2474 \\
CCC~\cite{keuper2018motion} & 51.2 & 75.9 & 20.9\% & 37.0\% & 25937 & 247822 & 1802 \\
FHFD~\cite{henschel2018fusion} & 51.3 & 77.0 & 21.4\% & \textbf{35.2\%} & 24101 & 247921 & 2648 \\
\midrule[0.5pt]
{\bf Ours} & \textbf{52.4} & 76.6 & \textbf{22.4\%} & 40.0\% & \textbf{20176} & \textbf{246158} & 2224 \\
\midrule[0.5pt]
\end{tabular}}}
\label{state-of-the-art-mot17}
\end{table*}

We compare the baseline tracking algorithm to the proposed point process enhanced tracking algorithm. The experiments are performed on MOT2016 with different components of our approach. The results are shown in \autoref{ablation_study}. To outline the superiority and generalization ability of the proposed method, we evaluated the point process model on other tracking methods~\cite{wojke2017simple}\cite{bergmann2019tracking}. \autoref{fig-5} provides visuals of some tracking results by the baseline tracking method and the point process enhanced tracking method. We also report the test AP and speed of point process with different components on MOT16 dataset, which is shown in \autoref{ablation_study_ap}. \hl{For reference, the speed of our baseline method is 59 FPS on the MOT17 testing data and when it is combined with Sync/Async RNN, the speed is 26 FPS.} \hl{It is worth noting that the entire procedure is actually a two-step process -- the point process method first removes ``bad'' detections in the sequence, and thereafter the remaining detections are used to construct the tracklets. As such, the point process does not affect the tracking speed. Therefore, we only compare the speed of the point process step using different components.}

\hl{In this comparison, the following tracking algorithms are evaluated upon to demonstrate the generalization ability of our approach.}

\noindent\textbf{Baseline:}
The tracklet based algorithm which is introduced in Section \ref{sec:tracklet}. We apply k-means clustering algorithm to generate reliable tracklets followed by softassign~\cite{wang2016joint} algorithm to associate the tracklets into trajectories. For simplicity, we use the terminal detection in each tracklet to calculate tracklet-wise similarity. The baseline algorithm includes a simple threshold-based strategy to filter out noisy detections with low confidences.

\noindent\textbf{Deep-SORT}~\cite{wojke2017simple}\textbf{:} Simple Online and Realtime Tracking (SORT) is a pragmatic approach to MOT with a focus on a simple yet effective algorithm. The deep-SORT variant integrates appearance information to improve the performance of SORT. It learns a deep association metric on large-scale person re-identification dataset in the offline pre-training stage. Then, during online application, it establishes measurement-to-track associations using nearest neighbor queries in the visual appearance space.

\noindent\textbf{Tracktor}~\cite{bergmann2019tracking}\textbf{:}
Tracktor is a tracker without any training or optimization on tracking data. It exploits the bounding box regression of an object detector to predict the position of an object in the next frame. It has good extensibility and performs well on three multi-object tracking benchmarks by extending it with a straightforward re-identification and camera motion compensation.

\hl{The following components are evaluated to demonstrate the effectiveness of our approach.}

\noindent\textbf{Time Independent:}
In this method, we only use the weight-shared CNN to handle the time series input. In other words, at each step, the current frame is input into a fully convolutional neural network (FCN)~\cite{long2015fully} to generate the intensity map for event prediction.
Note that this method has a ``memoryless'' mechanism, which does not capture any temporal relations within the input data or between event sequences.

\noindent\textbf{Synchronous RNN:}
This method, which is the partial version of our proposed method, takes only the time series as input data in order to reflect the state of environment where events happen. Correspondingly, the synchronous RNN models the exogenous intensity in our point process.

\noindent\textbf{Synchronous + Asynchronous RNN:}
This is the full version of our proposed method. It includes both synchronous and asynchronous RNN, where the time series and associated event sequence are taken as input data. We adopt a neural network based time evolving mechanism to align and merge these two features, for the generation of the intensity maps.

From \autoref{ablation_study}, \autoref{ablation_study_ap} and \autoref{fig-5}, we make the following observations:

(1) The time independent prediction method outperforms our baseline tracking algorithm. Tracking examples in \autoref{fig-5} show that directly associating the detected objects or simply using general tracklet association algorithms is ineffective and easily interfered by noisy or confusing detections. Comparatively, by applying the time independent model to predict and discard the ``bad'' detections before association process, we are able to mitigate these noisy and confusing detections to a reasonable measure of success.

(2) The Synchronous RNN model has significantly better results compared to the time independent model. The performance gains come from the fact that a learnable model of larger capacity is employed to capture the historical information to cater for complex temporal dynamics. In addition, we observe that both the time independent model and synchronous RNN model have lower MOTP values than the baseline algorithm; MOTP measures the precision of size and location of bounding boxes in trajectories. This is likely because these two methods mask out the ``bad'' detections effectively, so that they can improve tracking accuracy such as MOTA. This occurs at the expense of losing some ``good'' detections to mis-classification and masking. These errors occur sparsely in time, and can be recovered in the tracklet clustering process using linear interpolation (a common scheme also adopted by our baseline tracking algorithm). The only caution being that, bounding boxes generated by linear interpolation are less precise than the detector results especially when the objects exhibit large or highly dramatic motions. Nevertheless, we observe that the overlaps between interpolated boxes and ground-truth boxes are still larger than 50\%, which may lower the MOTP score, but will not affect the MOTA score.

(3) The full version of the proposed method (synchronous + asynchronous RNN) performs much better than all other compared methods. This demonstrates the importance of adopting event sequence in addition to time series as inputs. Intuitively, this helps the model to better capture long-term dependencies of events. From another aspect, the asynchronous RNN also reflects the endogenous intensity, as a complementary part to the exogenous intensity already modeled by synchronous RNN. Besides, \autoref{ablation_study_ap} also demonstrates the effectiveness of proposed method in predicting the ``bad'' detection results.

(4) Our proposed method consistently improve all the three methods, including the baseline algorithm, the alternative Deep-SORT~\cite{wojke2017simple} tracking algorithm and the Tracktor~\cite{bergmann2019tracking} algorithm in all metrics, which demonstrates its effectiveness and strong generalization ability.

\subsection{Comparison with State-of-the-art Methods}
\begin{figure}[t]
\small
\centering
\includegraphics[width=1\linewidth]{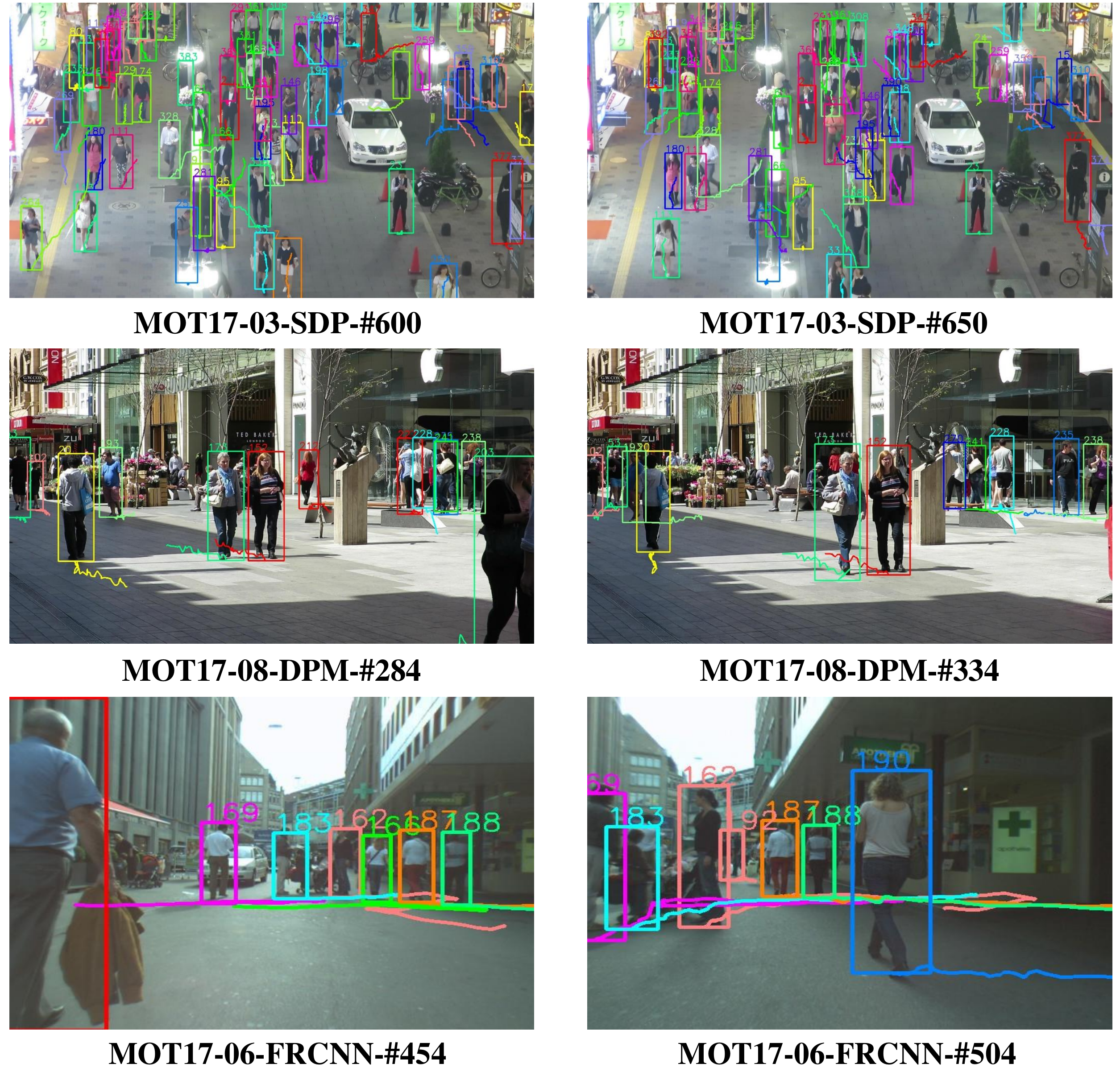}
\vspace{-4.5mm}
  \caption{Some tracking results of our method on MOT17 dataset.}
\vspace{-1.0ex}
\label{fig-6}
\end{figure}

With the best settings of the proposed method affirmed in the ablation study, we conduct further comparisons against state-of-the-art MOT tracker on the widely used benchmark dataset: MOT2016 and MOT2017. For fair comparison, all the methods are evaluated and reported based on the same evaluation protocol and metrics. 
\autoref{state-of-the-art-mot16} and \autoref{state-of-the-art-mot17} list the benchmark results for MOT2016 and MOT2017 respectively, comparing the proposed method against recent state-of-the-art MOT trackers such as OICF~\cite{kieritz2016online}, CBDA~\cite{bae2017confidence}, QCNN~\cite{son2017multi}, STAM~\cite{chu2017online}, MDM~\cite{tang2016multi}, NOMT~\cite{choi2015near}, JGD-NL~\cite{levinkov2017joint}, TSN-CC~\cite{peng2018tracklet}, LM-PR~\cite{tang2017multiple}, EEBMM~\cite{maksai2019eliminating}, NG-bL~\cite{kim2018multi}, OGSDL~\cite{fu2018particle}, DMAN~\cite{zhu2018online}, EDM~\cite{chen2017enhancing}, MHT~\cite{kim2015multiple}, DLCS~\cite{long2018real}, CCC~\cite{keuper2018motion}, and FHFD~\cite{henschel2018fusion}. \autoref{fig-6} illustrates some tracking results of our method on MOT17 dataset.

From the tracking performance comparisons in \autoref{state-of-the-art-mot16} and \autoref{state-of-the-art-mot17}, it can be seen that our tracker surpasses all other methods in terms of the primary evaluation metric MOTA. In MOT16, compared to the closest competitor LM-PR~\cite{tang2017multiple}, our method achieves a 1.7\% improvement (50.5\% \textit{vs.} 48.8\%) in MOTA, a 1.4\% improvement (19.6\% \textit{v.s.} 18.2\%) in MT and a 0.7\% improvement (39.4\% \textit{v.s.} 40.1\%) in ML. In MOT17, compared to the closest competitor FHFD~\cite{henschel2018fusion}, our method achieves a 1.1\% improvement (52.4\% \textit{vs.} 51.3\%) in MOTA and a 1.0\% improvement (22.4\% \textit{vs.} 21.4\%) in MT. Note that both LM-PR~\cite{tang2017multiple} and FHFD~\cite{henschel2018fusion} have more sophisticated and computationally heavy tracking pipelines than our baseline algorithm. This implies that we can make significant strides to improve the state-of-the-art by formulating it as a point process method.
On the other hand, the MOTP of our approach is slightly lower than some methods because the interpolated detections tend to be less precise when the objects have some large or highly dramatic motions.

Our approach also produces the lowest FN on both MOT16/MOT17 and highest MT in MOT17 (second highest on MOT16), which shows that the proposed method can accurately associate the tracklets and the interpolation process is effectively in generating missing detections. 
Our approach produces the lowest FP on MOT17 (second lowest on MOT16), which demonstrates the strengths of our approach in handling noisy and confusing detections by formulating these ``bad'' detections using our spatio-temporal point process model.

\textcolor{black}{
Although some methods (e.g., CCC~\cite{keuper2018motion}) have better performances on the ML and IDS metrics, they obtain this at the expense of sacrificing the performance on other metrics (i.e., MT and FP), which leads to a low overall result in MOTA. Comparatively, our approach can comprehensively balance different aspects in tracking and obtain a better overall result in MOTA. Besides, our approach has relatively high values on some metrics (i.e., IDS) because we use simple features and association schemes to perform tracking. In fact, since our proposed point-process model is generic, we can easily incorporate into our model more sophisticated features and association schemes (e.g., aggregated local flow descriptor in NOMT~\cite{choi2015near}) to potentially obtain better performances on these metrics.}
\section{Conclusion}
In this paper, we address the issue of mis-detections in Multiple Object Tracking (MOT) task by proposing a novel framework that effectively predicts and masks out noisy and confusing detection results prior to associating objects into trajectories. We formulate the ``bad'' detection results as a sequence of events, adopting the spatio-temporal point process to model such event sequences. A convolutional recurrent neural network (Conv-RNN) is introduced to instantiate the point process, where the temporal and spatial evolutions are well captured. Experimental results demonstrate notable performance improvement with the proposed method over state-of-the-art MOT algorithms across different metrics and benchmark datasets.

\bibliographystyle{IEEEtran}
\bibliography{egbib}

\end{document}